\documentclass[pdflatex]{jors}

\usepackage[numbers]{natbib} % added KB
\usepackage{graphicx} % added KB
\usepackage{listings} % added by KB
\usepackage{xcolor}% added by KB
\usepackage{hyperref}
\definecolor{codegreen}{rgb}{0,0.6,0}% added by KB
\definecolor{codegray}{rgb}{0.5,0.5,0.5}% added by KB
\definecolor{codepurple}{rgb}{0.58,0,0.82}% added by KB
\definecolor{backcolour}{rgb}{0.95,0.95,0.92}% added by KB

\lstdefinestyle{mystyle}{
    backgroundcolor=\color{backcolour},   
    commentstyle=\color{codegreen},
    keywordstyle=\color{magenta},
    numberstyle=\tiny\color{codegray},
    stringstyle=\color{codepurple},
    basicstyle=\ttfamily\footnotesize,
    breakatwhitespace=false,         
    breaklines=true,                 
    captionpos=b,                    
    keepspaces=true,                 
    numbers=left,                    
    numbersep=5pt,                  
    showspaces=false,                
    showstringspaces=false,
    showtabs=false,                  
    tabsize=2
} % added by KB

\font\myfont=cmr12 at 17pt

\begin{document}

\title{\myfont arfpy: A python package for density estimation and generative modeling with adversarial random forests}
\author{\textbf{Kristin Blesch}$^{*1,2}$ \and \textbf{Marvin N. Wright}$^{1,2,3}$}
\date{%
   \small $^*$ corresponding author; blesch@leibniz-bips.de\\
   $^1$Leibniz Institute for Prevention Research and Epidemiology -- BIPS, Germany; \\%
    $^2$Faculty of Mathematics and Computer Science, University of Bremen, Germany; \\
    $^3$Department of Public Health, University of Copenhagen, Denmark%
}
\maketitle

% \author[1,2,3]{\fnm{Marvin N.} \sur{Wright}}\email{wright@leibniz-bips.de}

% \affil*[1]{\orgname{Leibniz Institute for Prevention Research \& Epidemiology – BIPS}, \orgaddress{\city{Bremen}, \country{Germany}}}

% \affil[2]{\orgdiv{Faculty of Mathematics and Computer Science}, \orgname{University of Bremen}, \orgaddress{ \city{Bremen}, \country{Germany}}}

% \affil[3]{\orgdiv{Department of Public Health}, \orgname{University of Copenhagen}, \orgaddress{\city{Copenhagen}, \country{Denmark}}}

\section*{Abstract}

\textcolor{black}{
%A short (ca. 100 word) summary of the software being described: what problem the software addresses, how it was implemented and architected, where it is stored, and its reuse potential.
This paper introduces \texttt{arfpy}, a python implementation of Adversarial Random Forests (ARF) (Watson et al., 2023), which is a lightweight procedure for synthesizing new data that resembles some given data. The software \texttt{arfpy} equips practitioners with straightforward functionalities for both density estimation and generative modeling. The method is particularly useful for tabular data and its competitive performance is demonstrated in previous literature. As a major advantage over the mostly deep learning based alternatives, \texttt{arfpy} combines the method's reduced requirements in tuning efforts and computational resources with a user-friendly python interface. This supplies audiences across scientific fields with software to generate data effortlessly. \\ \url{https://github.com/bips-hb/arfpy}
}

\section*{Keywords}

\textcolor{black}{Generative Modeling; Density Estimation; Machine Learning}

\section*{Introduction}

\textcolor{black}{
Generative modeling is a challenging task in machine learning that aims to synthesize new data which is similar to a set of given data. State of the art are computationally intense and tuning-heavy algorithms such as generative adversarial networks (GANs)\citep{goodfellow2014, xu2019}, variational autoencoders \citep{kingma2014}, normalizing flows \citep{rezende2015}, diffusion models \citep{ho2020} or transformer-based models \citep{vaswani2017}. 
A much more lightweight procedure is to use an Adversarial Random Forest (ARF) \citep{watson2023}.  ARFs achieve competitive performance in generative modeling with much faster runtime \citep{watson2023}, yet they do not require the practitioner to have extensive knowledge of generative modeling.\\
Further, ARFs are especially useful for data that comes in a table format, i.e., tabular data. That is because ARFs are based on random forests \citep{breiman2001} which leverage the advantages that tree-based methods have over neural networks on tabular data \citep{grinsztajn2022} for generative modeling. Further, as part of the procedure, ARFs give access to the estimated joint density, which is useful for several other fields of research, e.g., unsupervised machine learning. For the task of density estimation, ARFs have been demonstrated to yield remarkable results as well \citep{watson2023}.  In brief, ARFs are a promising methodological contribution to the field of generative modeling and density estimation, providing a ready-made solution to generate data for practitioners across fields.\\
ARFs have already gained some attention in the scientific community \citep{nock2023}, however, the paper by \cite{watson2023} provides the audience with a R software package only. The machine learning and generative modeling community, however, is mostly using python as a programming language and to reach a broad audience more generally, a fast and user-friendly implementation of ARFs in python is highly desirable. We aim to fill this gap with the presented python implementation of ARFs. \\
\texttt{arfpy} is inspired by the R implementation called \textit{arf} \citep{wright2023}, but transfers the algorithmic structure to match the class-based structure of python code and takes advantage of computationally efficient python functions. Similar to the R implementation, separate functions for first fitting the density (FORDE algorithm \citep{watson2023}) and then generating new data samples (FORGE algorithm \citep{watson2023}) exist. However, in \texttt{arfpy}, the functions are called for an initialized ARF class object, which is showcased in the usage example below.\\
Crucially, for practitioners working with python as programming language, the direct python implementation is more robust and convenient to users than calling fragile wrappers like \texttt{rpy2} \citep{rpy2} that aim to make R code running in python. The benefits of a direct python implementation of ARFs for the generative modeling community have already been recognized by now. For example, \texttt{arfpy} is integrated in the data synthesizing framework \texttt{synthcity} \citep{synthcity}.
}

\section*{Implementation and architecture}
\begin{figure}[ht] 
	\centering
	\includegraphics[width=0.7\textwidth]{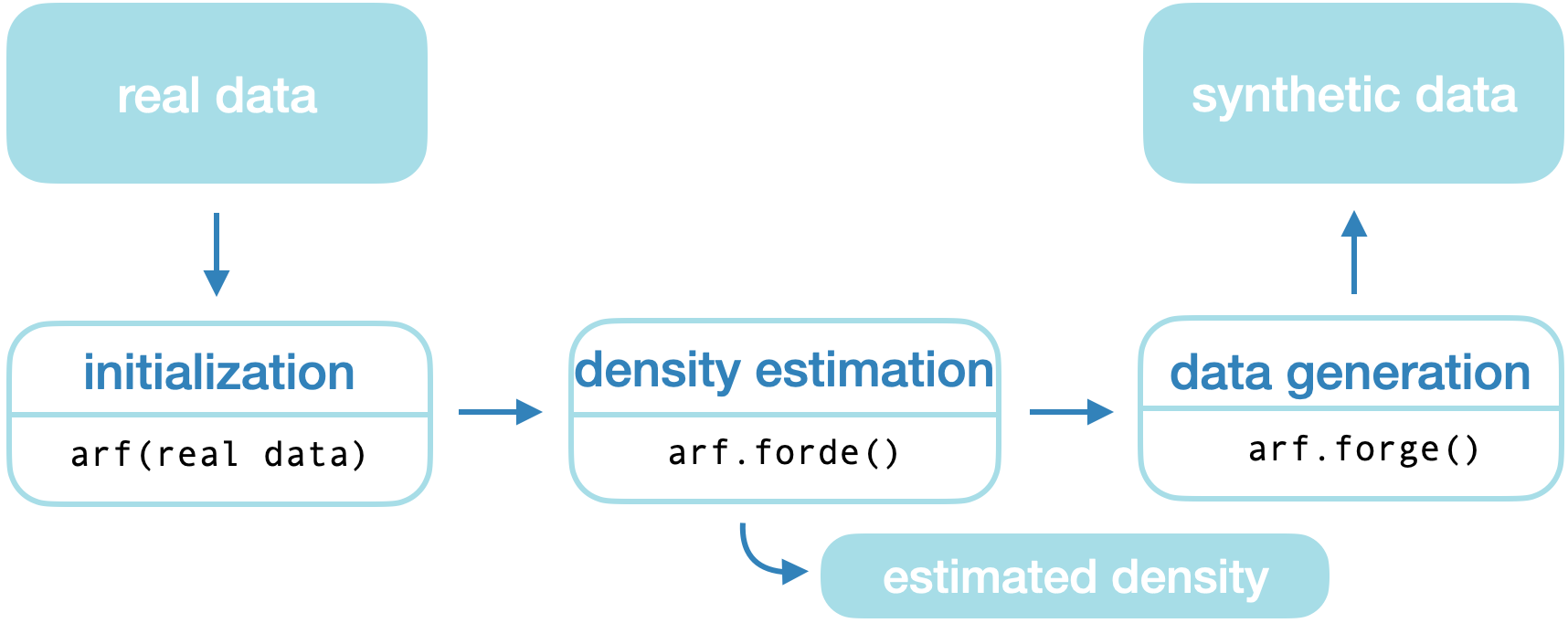}
	\caption{Workflow of \texttt{arfpy}'s core functionalities.}
	\label{fig::workflow}
\end{figure}

\paragraph{Module Design}
\textcolor{black}{
The general workflow of generating data with \texttt{arfpy} is (1) to initialize an object of class \texttt{arf} with real data, (2) estimate the density and (3) sample new data. This procedure is visualized in Figure \ref{fig::workflow}. \\
The architecture of \texttt{arfpy} reflects this workflow and we have class \texttt{arf} building the backbone of the procedure. An instance of class \texttt{arf} takes the real data set as input and trains an ARF, i.e., learns the actual data's structure. To this object, functions to estimate the density (FORDE algorithm \citep{watson2023}, function \texttt{forde()}) and generate data (FORGE algorithm \citep{watson2023}, function \texttt{forge()}) can be applied. This architecture allows users to learn the structure of the real data once (when initializing the \texttt{arf} class object) and then flexibly adapt density estimation, e.g., using different parameters, or repeatedly sampling new data without having to refit the model.}

\paragraph{Methodology Overview}
\textcolor{black}{For interested readers, we want to briefly describe the methodological foundations of ARFs, but refer to \citep{watson2023} for further details. From a given real data set, first, naive synthetic data is generated (initial generation step) by sampling from the marginal distributions of the features.  Then, a random forest \citep{breiman2001} is fit to distinguish this synthetic from the real data (initial discrimination step). This procedure, also known as fitting an unsupervised random forest \citep{shi2006}, guides the random forest to learn the dependency structure in the data. Using this forest, we can sample observations from the leaves of the trees to generate updated synthetic data (generation step). Subsequently, a new random forest is fit to differentiate between synthetic and real data (discrimination step). Drawing on the adversarial idea of GANs, this iterative procedure of data generation and discrimination will be repeated until the discriminator cannot distinguish between generated and real data anymore. At this stage, the accuracy of the forest will be $\leq 0.5$ and the forest is assumed to have converged, which implies mutually independent features in the terminal nodes. This drastically simplifies density estimation and generative modeling, as it allows us to formulate the univariate density for each feature separately with data in the leaves of the fitted ARF (FORDE algorithm) and then combine them to the joint density, instead of having to model multivariate densities. For data generation, we can use this trait to sample a new observation by drawing a leaf from the forest of the last iteration step and use the data distributions with parameters estimated from that leaf to sample each feature separately (FORGE algorithm).}

\begin{figure}[ht] 
	\centering
	\includegraphics[width=1.1\textwidth, trim={5cm 25cm 0 0},clip]{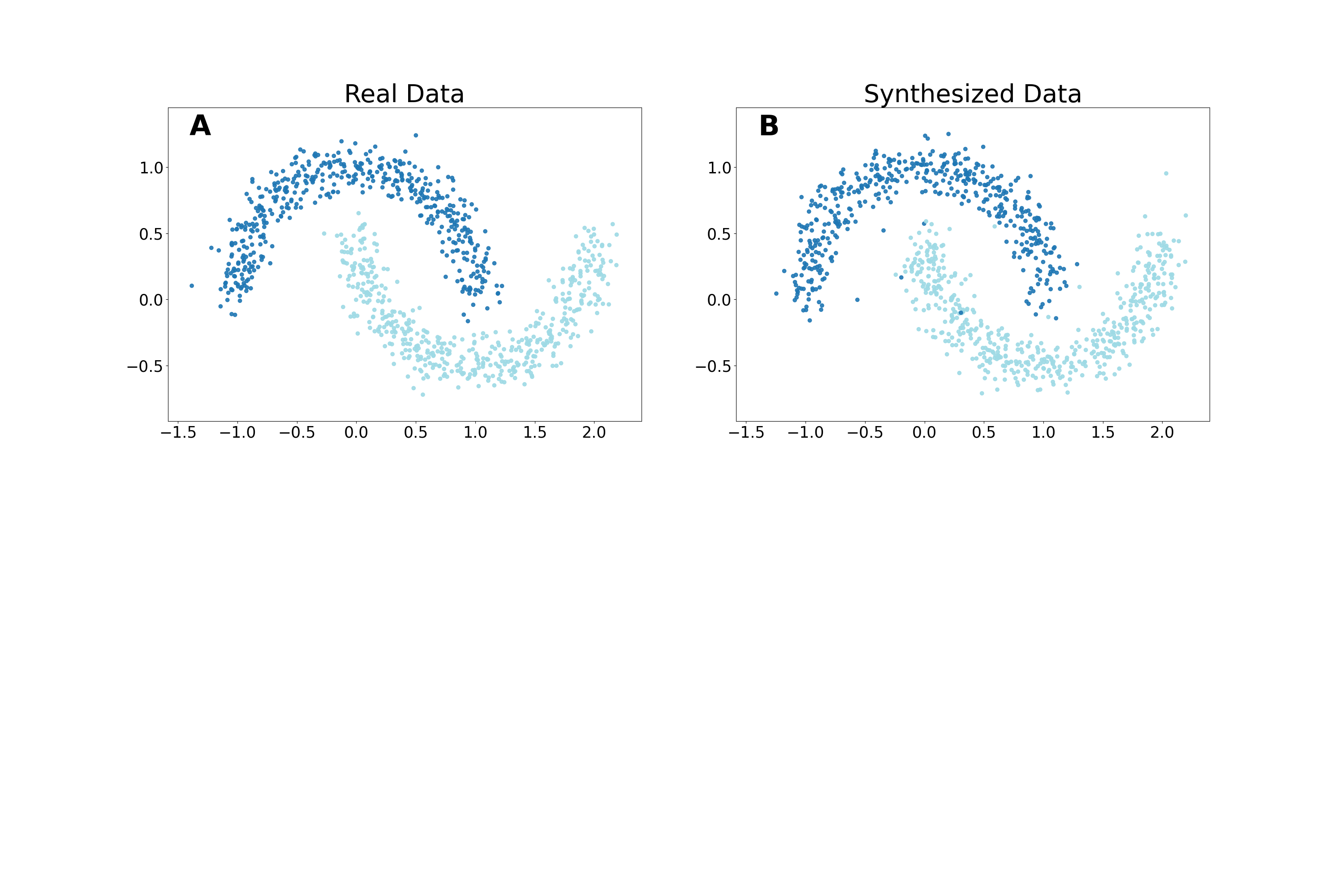}
	\caption{Comparison of real and synthesized data. }
	\label{fig::twomoons}
\end{figure}

\paragraph{Example Usage}
\textcolor{black}{Let us illustrate the usage of \texttt{arfpy} with a visually intuitive example: We create data using \texttt{make\_moons} from \texttt{sklearn.datasets}, which results in data along two continuous axes that looks like two moons from different categories. Statistically speaking, this is a tabular dataset, consisting of both continuous and categorical features that exhibit a dependency structure. For a more intuitive understanding of the data, see Figure \ref{fig::twomoons}, Panel \textbf{A}. The task of \texttt{arfpy} is to learn the structure of this given (real) data and generate new data instances that appear similar.} 

\textcolor{black}{To initialize the workflow, we need to run relevant imports, including the import of class \texttt{arf} from the \texttt{arfpy} module, and create the real dataset. The \texttt{arf} class takes a \texttt{pandas DataFrame} as input, so the real data is pre-processed to match this requirement. This incorporates setting unique column names (\texttt{'dim\_1', 'dim\_2','label'}) and ensuring that \texttt{'label'} is stored in the correct data type \texttt{'category'}.} 

\begin{lstlisting}
import pandas as pd
from sklearn.datasets import make_moons
from arfpy import arf

moons_X, moons_y = make_moons(n_samples = 3000,  noise=0.1)  
df = pd.DataFrame({"dim_1" : moons_X[:,0], "dim_2" : moons_X[:,1], "label" : moons_y})
df['label'] = df['label'].astype('category')

df.head()

#>   dim_1	     dim_2	        label
#>	 1.782717	   0.099124	      1
#>	 1.087497	   0.298744	      0
#>  -0.576695	   0.801675	      0
#>	 0.623931   -0.506896	      1
\end{lstlisting}

\textcolor{black}{With the real dataset preprocessed as needed, we can proceed with training the ARF to learn the data's structure. Creating an object of class \texttt{arf} will trigger ARF model fitting using the data provided. }

\begin{lstlisting}
my_arf = arf.arf(x = df) 

#> Initial accuracy is 0.82
#> Iteration number 1 reached accuracy of 0.36
\end{lstlisting}

\textcolor{black}{Because we have used the parameter default \texttt{verbose = True}, the training of \texttt{my\_arf} prints out some interesting information: The initial accuracy, which corresponds to the accuracy of the random forest in distinguishing real data from naive synthetic data, is $0.82$. This implies that the random forest is doing very well in distinguishing real from naive synthetic data and therefore, we can assume the model to have learned relevant dependencies that allow the model to make this distinction. Using this forest to sample updated synthetic data, and fitting a new random forest to distinguish this data from real data leads to an accuracy of only $0.36$. This accuracy is below the default threshold of $0.5$, so loosely speaking, the synthetic data generated with the forest cannot be accurately distinguished from real data, i.e., the generated data looks like real data, which is the goal the algorithm was aiming for. In other words, the relevant dependency structures of the real data have been learned by the forest in the first round of iteration already, so the algorithm has converged and no further iterations need to be conducted.} 

\textcolor{black}{After the ARF has converged, we can proceed to estimating the joint density. Recap that in a converged ARF, the features are mutually independent in the leaves, which simplifies the challenging multivariate density estimation task into many simple univariate density estimation tasks. The joint density is then a factorization of the individual density estimates across leaves in the ARF. We can call function \texttt{forde()} on the \texttt{my\_arf} object to estimate the density and store the returned dictionary to explore the parameters.
The \texttt{FORDE} dictionary contains the estimated parameters for continuous (key \texttt{'cnt'}) and categorical features (key \texttt{'cat'}). As mentioned in the above paragraph, the parameters are estimated using the data points in the forest's leaves, so we will get estimates for each leaf individually. The parameters for the categorical features simply correspond to the empirical frequency of categories in the leaves, so for a more complex example, we can take a look at the continuous feature's parameter estimates in  \texttt{FORDE['cnt']}. We have used the default distribution (truncated normal distribution) to model continuous features, so the output will reflect estimates for the mean and standard deviation for each feature (\texttt{'dim\_1'}, \texttt{'dim\_2'}) in each leaf, which is uniquely identified by \texttt{'tree'} and \texttt{'nodeid'}:}
\\
\begin{lstlisting}
FORDE = my_arf.forde()

FORDE['cnt'].iloc[:,:5].head()

#>   tree	nodeid	 variable	  mean	    sd
#>   0	  3	       dim_1	    0.961437	0.214925
#>	 0	  3	       dim_2	   -0.671571	0.028193
#>   0	  11	     dim_1	    1.040565	0.185581
#>	 0	  11	     dim_2	   -0.621924	0.003328

\end{lstlisting}

\textcolor{black}{With the parameters estimated, we can move on to the final step of the generative modeling task and sample new data instances with the function \texttt{forge()}. \\
For each instance to be generated, the function randomly samples a leaf from the forest with weighted probability according to the coverage of real data in the leaves of the ARF and then uses the parameters estimated through \texttt{forde()} to sample a new data instance. 
}

\begin{lstlisting}
df_syn = my_arf.forge(n = 1000)

df_syn.head()

#>   dim_1	     dim_2	        label
#>	-0.018004	   0.283963	      1
#>	 1.734200	  -0.085115	      1
#>	-0.009840	   1.046872	      0
#>	 0.868400	  -0.352692	      1

\end{lstlisting}

\textcolor{black}{Calling \texttt{forge()} completes the task of generating synthetic data that mimics real data. From the generated data table itself, the similarity is hard to grasp, but we can visually inspect the quality of the synthetic data in Figure \ref{fig::twomoons}.}

\section*{Quality control}

\textcolor{black}{
The software has been tested through unit tests, which includes testing of relevant functionalities with various input data sets. The workflow of running these tests is automated on GitHub actions, but can be run locally and with customized data sets using the instructions provided in the software repository.
Further, the repository allows to publicly raise questions or report bugs and gives clear guidelines on how to contribute to the open source software project are lined out.}

\section*{(2) Availability}
\vspace{0.5cm}
\section*{Operating system}

\textcolor{black}{Platform Independent}

\section*{Programming language}

 \textcolor{black}{Python  $\ge$ 3.8}

\section*{Additional system requirements}

\textcolor{black}{No specific requirements}

\section*{Dependencies}

\textcolor{black}{numpy $\ge$ 1.20.3, pandas $\ge$ 1.5, scikit-learn $\ge$ 0.24, scipy $\ge$ 1.4}

\section*{List of contributors}

\textcolor{black}{Blesch, Kristin$^{a, b}$;  \\
Wright, Marvin N.$^{a, b, c}$; \\
(a) Leibniz Institute for Prevention Research and Epidemiology -- BIPS, Germany; \\
(b) Faculty of Mathematics and Computer Science, University of Bremen, Germany; \\
(c) Department of Public Health, University of Copenhagen, Denmark
}

\section*{Software location:}

{\bf Archive} \textcolor{black}{} 

\begin{description}[noitemsep,topsep=0pt]
	\item[Name:] \textcolor{black}{arfpy}
	\item[Persistent identifier:] \textcolor{black}{\url{https://pypi.org/project/arfpy/}}
	\item[Licence:] \textcolor{black}{MIT}
	\item[Publisher:]  \textcolor{black}{Kristin Blesch}
	\item[Version published:] \textcolor{black}{v0.1.1}
	\item[Date published:] \textcolor{black}{22/09/2023}
\end{description}

{\bf Code repository} \textcolor{black}{}

\begin{description}[noitemsep,topsep=0pt]
	\item[Name:] \textcolor{black}{arfpy}
	\item[Persistent identifier:] \textcolor{black}{\url{https://github.com/bips-hb/arfpy}}
	\item[Licence:] \textcolor{black}{MIT}
	\item[Date published:] \textcolor{black}{06/09/2023}
\end{description}

\section*{Language}

\textcolor{black}{English}

\section*{(3) Reuse potential}

\textcolor{black}{
ARFs have been introduced with a solid theoretical background, yet do not have to compromise on a complex algorithmic structure and instead are a low-key algorithm that does not require extensive hyperparameter tuning \citep{watson2023}. In contrast to the typically deep learning based alternatives, ARF does not require background knowledge of generative modeling, intense tuning efforts or large computational resources. Given the theoretical foundation and straightforward implementation with \texttt{arfpy}, the methodology is attractive for both scholars conducting rather theoretical research in statistics, e.g., density estimation, as well as practitioners from other fields that need to generate new data samples. \\
Typical use cases of such synthesized data samples are, for example, the imputation of missing values, data augmentation or the conduct of analyses that respect data protection rules.
With the specialty of ARFs being particularly suitable for tabular data, including a natural incorporation of both continuous and categorical features, the straightforward python implementation of ARFs provides a convenient algorithm to a broad audience from different fields. \\ 
With the python programming language being widespread, \texttt{arfpy} can smoothly integrate in the code of various applications. Further, usability is enhanced by the intuitive documentation provided at \url{https://bips-hb.github.io/arfpy/}, making \texttt{arfpy} an easily accessible tool to generate data.  \\
In sum, \texttt{arfpy} introduces density estimation and generative modeling with ARFs to python, which enables practitioners from a wide variety of fields to generate fast and reliable synthetic data and density estimates with python as a programming language. }

\section*{Acknowledgements}

\textcolor{black}{We thank David S. Watson and Jan Kapar for their contributions to establishing the theoretical groundwork of adversarial random forests. }

\section*{Funding statement}

\textcolor{black}{This work was supported by the German Research Foundation (DFG), Emmy Noether Grant 437611051.}

\section*{Competing interests}

\textcolor{black}{The authors declare that they have no competing interests.}

%\section*{References}

%\textcolor{black}{Please enter references in the Harvard style and include a DOI where available, citing them in the text with a number in square brackets, e.g. \\ }
\bibliography{bib}

\vspace{2cm}

\end{document}